\documentclass[journal]{IEEEtran}

\usepackage{url}
\usepackage{stfloats}
\usepackage{multirow}
\usepackage{amsmath}
\usepackage{amsfonts}
\usepackage{algorithmic}
\usepackage{algorithm}
\usepackage{mathrsfs}
\usepackage{amssymb}
\usepackage{makecell}
\usepackage{tabularx}

\usepackage{graphicx}

\begin{document}

\title{AWEncoder: Adversarial Watermarking Pre-trained Encoders in Contrastive Learning}

\author{Tianxing Zhang, Hanzhou Wu, Xiaofeng Lu and Guangling Sun

\thanks{All the authors in this manuscript are from the School of Communication and Information Engineering, Shanghai University, Shanghai 200444, China; E-mails: zhang\_tx@shu.edu.cn, h.wu.phd@ieee.org, luxiaofeng@shu.edu.cn, sunguangling@shu.edu.cn. \emph{Corresponding author: Guangling Sun}}
\thanks{\emph{Hanzhou Wu improved and proofread this manuscript seriously.}}
}

\markboth{}
{}
\maketitle

\begin{abstract}
As a self-supervised learning paradigm, contrastive learning has been widely used to pre-train a powerful encoder as an effective feature extractor for various downstream tasks. This process requires numerous unlabeled training data and computational resources, which makes the pre-trained encoder become valuable intellectual property of the owner. However, the lack of a priori knowledge of downstream tasks makes it non-trivial to protect the intellectual property of the pre-trained encoder by applying conventional watermarking methods. To deal with this problem, in this paper, we introduce AWEncoder, an adversarial method for watermarking the pre-trained encoder in contrastive learning. First, as an adversarial perturbation, the watermark is generated by enforcing the training samples to be marked to deviate respective location and surround a randomly selected key image in the embedding space. Then, the watermark is embedded into the pre-trained encoder by further optimizing a joint loss function. As a result, the watermarked encoder not only performs very well for downstream tasks, but also enables us to verify its ownership by analyzing the discrepancy of output provided using the encoder as the backbone under both white-box and black-box conditions. Extensive experiments demonstrate that the proposed work enjoys pretty good effectiveness and robustness on different contrastive learning algorithms and downstream tasks, which has verified the superiority and applicability of the proposed work. 
\end{abstract}


\IEEEpeerreviewmaketitle

\section{Introduction}
\IEEEPARstart{T}{he} rapid development of deep learning (DL) has benefited a lot from a large number of diverse labeled datasets, which, however, simultaneously has become a factor hindering the further development of DL. It is due to the reason that the cost of collecting sufficient high-quality datasets with correct labels is prohibitively expensive. Fortunately, the emergence of self-supervised learning (SSL) \cite{paper1} to train a powerful encoder with the unlabeled samples can effectively overcome the above obstacle. Contrastive learning \cite{paper2, paper3}, as one mainstream self-supervised learning technique, has achieved great success in various tasks. For example, SimCLR \cite{paper4} and MoCo \cite{paper5} have already enabled the SSL encoders to outperform traditional supervised learning based encoders in several downstream tasks. However, training the encoders still requires lots of unlabeled data and consumes a great deal of computing resources. Many malicious users are likely to steal the pre-trained encoders to obtain illegal income and even bring serious security risks \cite{paper6, paper7}. Therefore, we need to find reliable solutions to protect the intellectual property of the pre-trained encoders. 

As a major technique to protect the intellectual property of DL models, model watermarking \cite{paper8, paper9, paper10, Wu:TCSVT} is widely studied recently. It can be roughly divided into two categories: \emph{white-box watermarking} and \emph{black-box watermarking}. The former \cite{paper11, paper12} embeds a watermark into the internal parameters \cite{paper13}, feature maps \cite{paper14} or structures \cite{paper15} of a DL model, which needs white-box access to the watermarked model. The latter \cite{paper16} usually uses backdooring \cite{paper17, paper18} or similar techniques such as adversarial attack \cite{paper19} to mark a model. The ownership can be verified by analyzing the classification results of the marked model in correspondence to a set of carefully crafted samples. Compared with white-box watermarking, black-box watermarking is more desirable for applications since in most cases the model owner has no access to the internal details of the target model. However, for protecting the aforementioned encoders, the lack of a priori knowledge of downstream tasks makes it quite difficult to craft the special samples for black-box watermarking. It indicates that we cannot simply extend the existing black-box watermarking schemes to the encoders. As a result, we urgently need to develop novel watermarking schemes specifically for the pre-trained encoders. 

In this paper, we introduce AWEncoder, a copyright protection method for the pre-trained encoders in contrastive learning via an adversarial watermark. In the proposed method, through optimizing an adversarial perturbation \cite{paper20}, we determine such a perturbation according to the pre-trained encoder that it can cluster the perturbed training samples around a randomly selected key image in the embedding space and can also be used as the secret watermark. The watermark is then embedded into the pre-trained encoder via contrastive learning by further optimizing a joint loss involving watermark embedding. In this way, even if the downstream task of the watermarked encoder is unknown to us, we can still verify its ownership under the black-box condition. Experimental results demonstrate that the ownership can be reliably verified under both white-box and black-box conditions, which is quite helpful for applications.

In summary, the main contributions of this paper include:

\begin{itemize}
	\item We propose a novel adversarial watermarking strategy for the pre-trained encoders in the embedding space, which is more effective for watermark verification compared with previous black-box watermarking methods. 
	\item Unlike conventional watermarking methods assuming that the downstream task of the watermarked model is same as the original one, the proposed method does not require the priori knowledge of the downstream task. As a result, the proposed method enables us to verify the ownership under white-box and stricter black-box settings, which is more applicable to practice compared with previous arts. 
	\item Extensive experimental results demonstrate that the proposed method has satisfactory ability to verify the ownership of the target model and resist common watermark removal attacks such as model fine-tuning \cite{paper21} and model pruning \cite{paper22}, which has good application prospects.
\end{itemize}

The rest structure is organized as follows. We first provide preliminaries in Section II, followed by the proposed method in Section III. Experimental results and analysis are provided in Section IV. Finally, we conclude this paper in Section V.

\section{Preliminaries}
\subsection{Contrastive Learning}
Contrastive learning is one of the most popular SSL techniques. SSL learns from unlabeled data, which can be regarded as an intermediate form between supervised and unsupervised learning. In this paper, we watermark a pre-trained encoder by contrastive learning. To this end, we consider two representative contrastive learning algorithms SimCLR \cite{paper4} and MoCo v2 \cite{paper23}. We briefly describe them in the following. It is pointed that the proposed method is not subject to this two algorithms. 

\subsubsection{SimCLR \cite{paper4}} SimCLR aims to learn representations by maximizing agreement between differently augmented views of the same sample via a contrastive loss function in the latent space. Briefly, SimCLR consists of three modules that are \emph{data augmentation}, \emph{feature encoder} $f$ and \emph{projection head} $g$. Data augmentation transforms a sample $\textbf{x}\in \mathcal{X}$ into two augmented views $\textbf{x}_i$ and $\textbf{x}_j$ treated as a positive pair. The feature encoder $f$ extracts representation vectors from augmented examples, e.g., $\textbf{h}_i = f(\textbf{x}_i)$ for the augmented sample $\textbf{x}_i$. The projection head $g$ maps representations to the space where the contrastive loss is applied, e.g., $\textbf{z}_i = g(\textbf{h}_i) = g(f(\textbf{x}_i))$. With $N > 0$ samples in a mini-batch, we are able to collect $2N$ augmented samples. Two augmented samples determined from the same sample constitute a positive pair. Otherwise, it is treated as a negative pair. Let $\text{sim}(\textbf{u}, \textbf{v}) = \textbf{u}^\text{T}\textbf{v}/||\textbf{u}||||\textbf{v}||$ represent the dot product between $\ell_2$ normalized $\textbf{u}$ and $\textbf{v}$ (i.e., cosine similarity). The loss function for a positive pair $(\textbf{z}_i, \textbf{z}_j)$ and a negative pair $(\textbf{z}_i, \textbf{z}_k)$ is then defined as \cite{paper4}:

\begin{equation}
\mathcal{L}_{i,j} = - \log \left(\frac{\exp \left(\operatorname{sim}\left(\mathbf{z}_{i}, \mathbf{z}_{j}\right) / \tau\right)}{\sum_{k=1}^{2 N} \mathbb{I}[k \neq i] \cdot \exp \left(\operatorname{sim}\left(\mathbf{z}_{i}, \mathbf{z}_{k}\right) / \tau\right)}\right),
\end{equation}
where $\tau$ denotes a temperature parameter and $\mathbb{I}[k\neq i] \in \{0, 1\}$ is an indicator function equal to 1 if and only if $k\neq i$.

\subsubsection{MoCo v2 \cite{paper23}} As an improved version of MoCo, MoCo v2 is consisting of a query encoder $f_q(\textbf{x};\theta _q)$ and a momentum key encoder $f_k(\textbf{x};\theta _k)$. The better performance of MoCo v2 is attributed to a large dictionary $\mathcal{K} = \{\textbf{k}_1, \textbf{k}_2, ..., \textbf{k}_{|\mathcal{K}|}\}$ which is the feature vectors of previous batches as the negative samples and it is continuously updating. A batch of $N$ samples will be encoded by $f_{q}$ as feature vectors and simultaneously encoded by $f_{k}$ as different augmentations. Suppose that there is a single key $\textbf{k}_+$ in the dictionary that an encoded query $\textbf{q}$ matches, the contrastive loss function is defined as \cite{paper23}:

\begin{equation}
\mathcal{L} = -\log \left(\frac{\exp \left(\mathbf{q}^\text{T} \mathbf{k}_{+} / \tau\right)}{\sum_{i=1}^{|\mathcal{K}|} \exp \left(\mathbf{q}^\text{T} \mathbf{k}_{i} / \tau\right)}\right).
\end{equation}
By minimizing the above contrastive loss function, the parameters of $f_q$, i.e., denoted by $\theta_q$, are updated by back-propagation, but the parameters of $f_k$, i.e., $\theta_k$, are updated by:
\begin{equation}
\theta_{k} \leftarrow \lambda \theta_{k} + (1 - \lambda) \theta_{q},
\end{equation}
where $\lambda \in [0, 1)$ is a momentum coefficient.

\subsection{Problem and Threats}
Mainstream black-box watermarking methods mainly focus on classification based models such as \cite{paper24, paper25}. The \emph{zero-bit} watermark is often embedded into the model by enforcing the model to learn the mapping relationship between the carefully crafted samples and the pre-determined labels. However, with the rise of contrastive learning, the pre-trained encoders are treated as feature extractors for various downstream tasks and the pre-training process is relying on the SSL strategy rather than label-based supervised learning \cite{paper26, paper27}. It indicates that traditional black-box watermarking techniques are not suitable for the pre-trained encoders in contrastive learning.

On the other hand, from the viewpoint of the adversary, he steals the encoder by an unauthorized way and add new layers to build the specific downstream task. The adversary may not train the encoder from scratch, but he is able to modify the encoder such as fine-tuning and pruning. In this case, even the owner of the encoder does not know what datasets and tasks will be used in downstream. As a result, when to watermark an encoder, it is very necessary for the embedded watermark to be transferable and robust. Moreover, it is quite desirable that the ownership of the encoder can be verified under both white-box condition and black-box condition. This has motivated the authors to propose a novel adversarial watermarking method.

\begin{figure*}[!t]
\begin{center}
\includegraphics[width=0.8\linewidth]{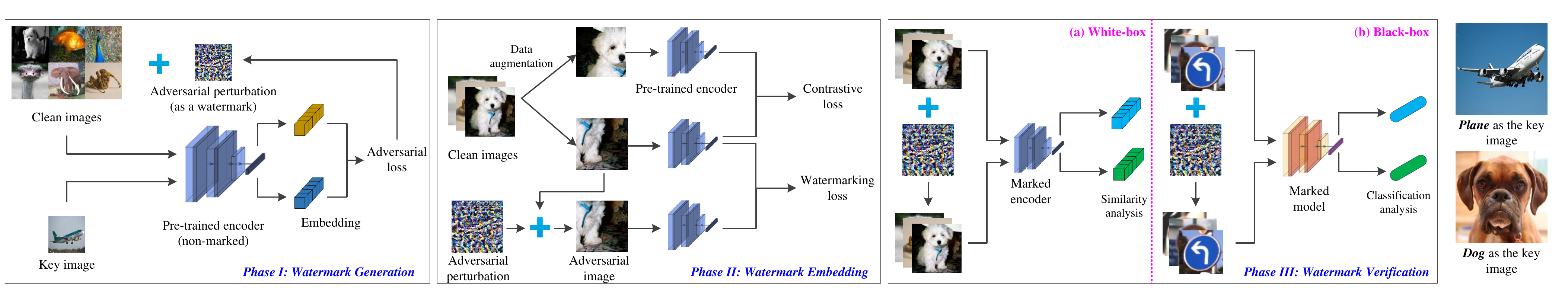}
\caption{General framework for the proposed method. We use the same dataset for Phase I, Phase II and Phase III (a) so that the watermark can be successfully embedded into $E_\theta$ while avoiding degrading the encoding performance of the encoder. In Phase II, each clean image is augmented into two images, only one randomly selected from which is used for generating the corresponding adversarial image that will be used for watermark embedding.}
\vspace{-0.5cm}
\end{center}
\end{figure*}

\section{Proposed Method}
Fig. 1 shows the general framework of the proposed method. In the following, we are to provide the technical details. 

\subsection{Watermark Generation}
The first step is to generate an adversarial perturbation $\textbf{w}_\text{adv}$ as the watermark using the pre-trained encoder $E_{\theta}$. Since our task is encoder oriented, the traditional perturbation optimization strategy based on classification labels is inapplicable. To deal with this problem, we use the embedding of a randomly selected image $\textbf{x}_\text{tar}$ (called \emph{key image}) extracted by $E_\theta$, i.e., $E_\theta(\textbf{x}_\text{tar})$, to generate $\textbf{w}_\text{adv}$ with a clean dataset $\mathcal{D}$. This enables the embedding of a perturbed image denoted by $E_\theta(\textbf{x}_i+\textbf{w}_\text{adv})$ (where $\textbf{x}_i \in \mathcal{D}, i \in \{1, 2, ..., |\mathcal{D}|\}$) cluster around $E_\theta(\textbf{x}_\text{tar})$. In other words, the distance between $E_\theta(\textbf{x}_\text{tar})$ and $E_\theta(\textbf{x}_i+\textbf{w}_\text{adv})$ is expected to be as low as possible. To achieve this goal, we minimize the following loss during perturbation optimization:
\begin{equation}
\mathcal{L}_\text{adv}= \mathbb{E}_{\textbf{x}_i\sim \mathcal{D}} \left [1-\text{sim}(E_\theta(\textbf{x}_i+\textbf{w}_\text{adv}), E_\theta(\textbf{x}_\text{tar})) \right ].
\end{equation}

By back-propagation, $\textbf{w}_\text{adv}$ can be generated without changing the parameters of $E_\theta$. It is obvious that different $\textbf{x}_\text{tar}$ result in different $\textbf{w}_\text{adv}$. By using $\textbf{x}_\text{tar}$ as a key, it is difficult for the adversary to forge the watermark. Some watermark generation methods by linearly superimposed fixed patterns \cite{paper28} may also perturb the embedding of an image, but their effect of changing the behavior of the encoder is weaker than our method.

\subsection{Watermark Embedding}
After generating $\textbf{w}_\text{adv}$, the next step is to embed $\textbf{w}_\text{adv}$ into the pre-trained $E_\theta$. It is realized by further training $E_\theta$ according to a combined loss $\mathcal{L}_\text{comb}$, which consists of two components, i.e., the contrastive loss $\mathcal{L}_\text{con}$ and the watermarking loss $\mathcal{L}_\text{wat}$. In other words, we have $\mathcal{L}_\text{comb} = \mathcal{L}_\text{con} + \alpha \mathcal{L}_\text{wat}$, where $\alpha$ is a parameter balancing the loss and we use $\alpha = 40$ by default. On one hand, $\mathcal{L}_\text{con}$ has already been defined in Section II, referring to Eq. (1) and Eq. (2). Notice that, $\mathcal{L}_\text{con}$ needs to be adjusted when other contrastive learning algorithms are applied. On the other hand, $\mathcal{L}_\text{wat}$ is defined as the KL divergence \cite{paper29} between the adversarial embedding and the non-adversarial embedding processed with the softmax function $\sigma$, i.e., 
\begin{equation}
\mathcal{L}_\text{wat} = \mathbb{E}_{\textbf{x}_i'\sim \mathcal{D}'}\left[\text{KL}(\sigma(E_\theta(\textbf{x}_i')), \sigma(E_\theta(\textbf{x}_i'+\textbf{w}_\text{adv})))\right],
\end{equation}
where $\textbf{x}_i'$ is sampled from the augmented dataset $\mathcal{D}'$. Thus, $E_\theta$ can be watermarked by updating its parameters during training. We will use $E_\theta^+$ to represent the marked version of $E_\theta$.  

\subsection{Watermark Verification}
When the defender verifies whether the suspicious encoder infringes the intellectual property of the watermarked encoder, white-box scenario and black-box scenario can be considered. In the white-box scenario, the defender has the access to the target encoder $E_\theta^- \approx E_\theta^+$. Namely, the defender can directly obtain the output of $E_\theta^-$ for watermark verification. We use the average KL divergence between a set of clean images and the corresponding adversarial images for similarity analysis:
\begin{equation}
T_\text{sim} = \frac{1}{|\mathcal{D}''|}\sum_{i=1}^{|\mathcal{D}''|}\text{KL}(\sigma(E_\theta^-(\textbf{x}_i'')), \sigma(E_\theta^-(\textbf{x}_i''+\textbf{w}_\text{adv}))),
\end{equation}
where $\mathcal{D}''$ is the clean dataset used for watermark verification. The watermark will be successfully verified if $T_\text{sim}$ is less than a threshold $t_s$, namely, we hope to keep $T_\text{sim}$ as low as possible. In our experiments, $|\mathcal{D}''|$ is set to 1000 by default.

In black-box scenario, given a suspicious downstream model $\mathcal{M}$, the defender wants to verify whether $\mathcal{M}$ is developed from $E_\theta^+$. The defender should build a clean dataset $\mathcal{D}^*$ related to the downstream task. Then, the classification performance for the downstream task is analyzed by:
\begin{equation}
T_\text{cls} = \frac{1}{|\mathcal{D}^*|}\sum_{i=1}^{|\mathcal{D}^*|}\mathbb{I}[\mathcal{M}(\textbf{x}_i^*)\neq  \mathcal{M}(\textbf{x}_i^*+\textbf{w}_\text{adv})].
\end{equation}
The watermark will be successfully verified if $T_\text{cls}$ is less than a threshold $t_c$, namely, we hope to keep $T_\text{cls}$ as low as possible. In our experiments, $|\mathcal{D}^*|$ is set to 1000 by default.

\emph{Remark:} Same as many existing adversarial attacking methods, $\textbf{w}_\text{adv}$ is generated by using back-propagation. The strength of $\textbf{w}_\text{adv}$ is constrained by a threshold $\epsilon $. To ensure that $\left \| \textbf{w}_\text{adv} \right \| _\infty \le \epsilon $, $\textbf{w}_\text{adv}$ is projected on the $\ell_\infty$ norm-ball around $\textbf{x}_i$ with radius $\epsilon$. We set $\ell_\infty$-norm and $\epsilon$ = 15 by default. With this perturbation strength, it can effectively cluster the perturbed images around the key image $\textbf{x}_\text{tar}$ in the embedding space. The implementation can be found in the released source code.

\section{Experimental Results and Analysis}
In our experiments, two benchmark datasets CIFAR-10 \cite{paper30} and ImageNet \cite{paper31} are used to pre-train the encoder. And, three benchmark datasets STL-10 \cite{paper32}, GTSRB \cite{paper33} and ImageNet are used for the downstream task of the encoder. For ImageNet, we randomly select 30 semantic categories out for training and another 10 categories different from training out for testing. For the watermark generation phase, the pre-trained encoders are ResNet-18 \cite{paper34} as the base model for SimCLR and MoCo v2 and the strength of the adversarial watermark is set as $\epsilon $ = 15. For the watermark embedding phase, we set the batch size to 50 for SimCLR and 32 for MoCo v2. Meanwhile, the total number of epochs is set to 50 and the learning rate is set to 0.003. Additionally, the key image is randomly selected from ImageNet which belongs to the semantic category ``Plane'' (see Fig. 1) and this image does not appear in the training set. We conduct simulation with PyTorch, accelerated by a single RTX 3080 GPU. To validate and reproduce our experiments, code will be released via \url{https://github.com/fc88zhang/AWEncoder}.

\begin{figure}[!t]
\begin{center}
\includegraphics[width=\linewidth]{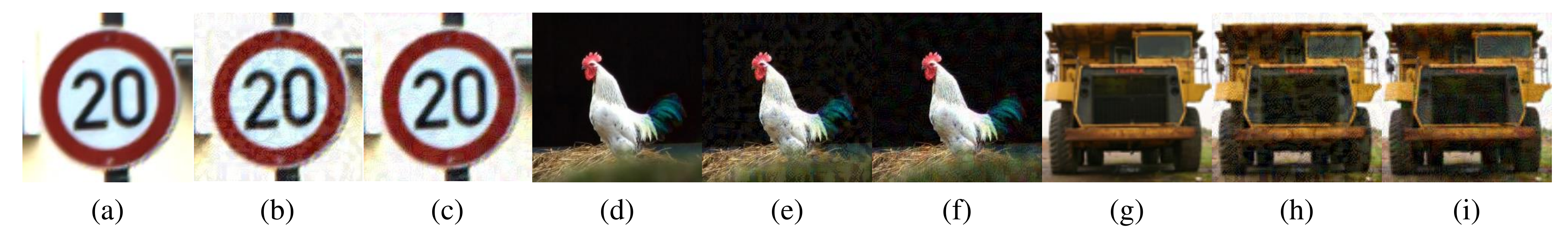}
\caption{Some examples for the adversarial image: (a, d, g) are clean images randomly selected from GTSRB, ImageNet and STL-10, respectively; (b, e, h) are adversarial with SimCLR; (c, f, i) are adversarial with MoCo v2.}
\vspace{-0.5cm}
\end{center}
\end{figure}

\begin{table}[!t]
\label{table1}
\centering
\caption{Effectiveness evaluation under the black-box condition.}
\resizebox{\columnwidth}{!}{
\begin{tabular}{c|c|c|c|cc|ccc}
\hline\hline
\begin{tabular}[c]{@{}c@{}}Pre-trained\\ dataset\end{tabular} & Encoder & \begin{tabular}[c]{@{}c@{}}Downstream\\ dataset\end{tabular} & Method & \multicolumn{2}{c|}{\begin{tabular}[c]{@{}c@{}}Accuracy\\ CE~/~WE\end{tabular}} & 
\multicolumn{3}{c}{\begin{tabular}[c]{@{}c@{}}$T_\text{cls}$\\ CE~/~WE~/~$|$CE-WE$|$\end{tabular}}
\\ \hline
\multirow{6}{*}{ImageNet} & \multirow{6}{*}{SimCLR} & \multirow{2}{*}{ImageNet} & WPE & 74.5\% & 73.4\% & 25.5\% & 86.4\% & 60.9\% \\
 &  &  & AWEncoder &  74.5\% & 71.1\% & 69.5\% & \textbf{3.9\%} & \textbf{65.6\%} \\ \cline{3-9} 
 &  & \multirow{2}{*}{GTSRB} & WPE & 77.8\% & 77.5\% & 31.7\% & 90.4\% & 58.7\% \\
 &  &  & AWEncoder & 77.8\% & 75.5\% & 83.2\% & \textbf{11.1\%} & \textbf{72.1\%} \\ \cline{3-9} 
 &  & \multirow{2}{*}{STL-10} & WPE & 64.7\% & 64.1\% & 27.3\% & 91.6\% & 64.3\% \\
 &  &  & AWEncoder &  64.7\% & 61.2\% & 86.3\% & \textbf{6.1\%} & \textbf{80.2\%}\\ \hline
\multirow{6}{*}{CIFAR-10} & \multirow{6}{*}{MoCo v2} & \multirow{2}{*}{ImageNet} & WPE & 72.3\% & 72.0\% & 21.3\% & 81.5\% & 60.2\% \\
 &  &  & AWEncoder &  72.3\% & 69.9\% & 67.6\% & \textbf{5.7\%} & \textbf{61.9\%}\\ \cline{3-9} 
 &  & \multirow{2}{*}{GTSRB} & WPE & 82.9\% & 80.7\% & 16.5\% & 84.6\% & 68.1\% \\
 &  &  & AWEncoder &  82.9\% & 82.7\% & 84.3\% & \textbf{10.9\%} & \textbf{73.4\%}\\ \cline{3-9} 
 &  & \multirow{2}{*}{STL-10} & WPE & 64.2\% & 63.9\% & 14.0\% & 82.8\%  & 68.8\%\\
 &  &  & AWEncoder &  64.2\% & 63.4\% & 80.5\% & \textbf{6.9\%} & \textbf{73.6\%} \\ \hline\hline
\end{tabular}}
\end{table}

\begin{table}[!t]
\label{table2}
\centering
\caption{Effectiveness evaluation under the white-box condition.}
\scalebox{0.65}{
\resizebox{\columnwidth}{!}{%
\begin{tabular}{c|c|c|c}
\hline\hline
\multicolumn{2}{c|}{\multirow{2}{*}{Model}} & \multicolumn{2}{c}{$T_\text{sim}$} \\ \cline{3-4} 
\multicolumn{2}{c|}{} & Correct watermark & Incorrect watermark \\ \hline
\multirow{2}{*}{SimCLR} & CE & 101.0 $\uparrow$ & 186.2 $\uparrow$ \\ 
& WE & \textbf{0.4} & 34.8 $\uparrow$  \\ \hline
\multirow{2}{*}{MoCo v2} & CE & 1.6 $\uparrow$ & 2.4 $\uparrow$ \\ 
& WE & \textbf{0.003} & 0.1 $\uparrow$ \\ \hline\hline
\end{tabular} }
}
\end{table}

\begin{table}[!t]
\label{table3}
\centering
\caption{Uniqueness evaluation due to different settings.}
\scalebox{0.65}{
\begin{tabular}{c|c|c|cc|cc}
\hline\hline
\multirow{2}{*}{\begin{tabular}[c]{@{}c@{}}Downstream\\ dataset\end{tabular}} & \multirow{2}{*}{Encoder} & \multirow{2}{*}{Setting} & \multicolumn{2}{c|}{SimCLR} & \multicolumn{2}{c}{MoCo v2} \\ \cline{4-7} 
 &  &  & \multicolumn{1}{c|}{$T_\text{cls}$} & $T_\text{sim}$ & \multicolumn{1}{c|}{$T_\text{cls}$} & $T_\text{sim}$ \\ \hline
\multirow{5}{*}{GTSRB} & \multirow{3}{*}{\begin{tabular}[c]{@{}c@{}}Pre-trained\\ encoder\end{tabular}} & \emph{Plane}, $\epsilon $ = 15 & \multicolumn{1}{c|}{\textbf{11.1\%}} & \textbf{0.4} & \multicolumn{1}{c|}{\textbf{10.9\%}} & \textbf{0.003} \\
 &  & \emph{Plane}, $\epsilon $ = 20 & \multicolumn{1}{c|}{80.6\% $\uparrow$} & 39.5 $\uparrow$ & \multicolumn{1}{c|}{72.9\% $\uparrow$} & 0.15 $\uparrow$ \\
 &  & \emph{Dog}, $\epsilon $ = 15 & \multicolumn{1}{c|}{70.6\% $\uparrow$} & 34.8 $\uparrow$ & \multicolumn{1}{c|}{68.4\% $\uparrow$} & 0.1 $\uparrow$ \\ \cline{2-7} 
 & \begin{tabular}[c]{@{}c@{}}Surrogate\\ encoder\end{tabular} & \emph{Plane}, $\epsilon $ = 15 & \multicolumn{1}{c|}{66.7\% $\uparrow$} & 22.9 $\uparrow$ & \multicolumn{1}{c|}{60.3\% $\uparrow$} & 0.04 $\uparrow$\\ \hline\hline
\end{tabular}%
}
\end{table}

\begin{center}
\begin{minipage}{\linewidth}
\begin{minipage}[t]{0.5\linewidth}
\makeatletter\def\@captype{table}
\setlength{\tabcolsep}{1.5mm}
\centering
\caption{Robustness against fine-tuning (under white-box condition).}
\setlength{\tabcolsep}{4mm}{
\scalebox{0.5}{
\label{table4}
\renewcommand{\arraystretch}{1.4} 
\begin{tabular}{c|cc|cc}
\hline\hline
\multirow{2}{*}{Fine-tuning} & \multicolumn{4}{c}{$T_\text{sim}$} \\ \cline{2-5} 
 & \multicolumn{2}{c|}{SimCLR} & \multicolumn{2}{c}{MoCo v2} \\ \hline
- & \multicolumn{2}{c|}{0.4} & \multicolumn{2}{c}{0.003} \\
FTAL & \multicolumn{2}{c|}{15.7} & \multicolumn{2}{c}{0.01} \\
RTAL & \multicolumn{2}{c|}{25.0} & \multicolumn{2}{c}{0.02} \\ 
\hline\hline
\end{tabular}
}}
\end{minipage}
\hspace{-0.10in}
\begin{minipage}[t]{0.5\linewidth}
\makeatletter\def\@captype{table}
\setlength{\tabcolsep}{1.5mm}
\centering
\caption{Robustness against pruning (under white-box condition).}
\setlength{\tabcolsep}{4mm}{
\scalebox{0.5}{
\label{table5}
\begin{tabular}{c|cc|cc}
\hline\hline
\multirow{2}{*}{Pruning ratio} & \multicolumn{4}{c}{$T_\text{sim}$} \\ \cline{2-5} 
 & \multicolumn{2}{c|}{SimCLR} & \multicolumn{2}{c}{MoCo v2} \\ \hline
- & \multicolumn{2}{c|}{0.40} & \multicolumn{2}{c}{0.003} \\
0.2 & \multicolumn{2}{c|}{0.66} & \multicolumn{2}{c}{0.005} \\
0.4 & \multicolumn{2}{c|}{0.66} & \multicolumn{2}{c}{0.008} \\
0.6 & \multicolumn{2}{c|}{0.68} & \multicolumn{2}{c}{0.01} \\
0.8 & \multicolumn{2}{c|}{0.72} & \multicolumn{2}{c}{0.04} \\

\hline\hline
\end{tabular}%
}}
\end{minipage}
\end{minipage}
\end{center}

\subsection{Effectiveness}
We first provide examples for the adversarial image in Fig. 1 Phase III. As shown in Fig. 2, due to the diversity, different images result in different degradation of the adversarial image, but the visual quality are all satisfactory, verifying the applicability of adversarial perturbation. It is admitted that one may use other perturbation strategy which is not our main focus. 

Then, we evaluate the effectiveness of watermarked encoder in both black-box and white-box scenarios. The encoders are pre-trained on two kinds of training datasets and contrastive learning algorithms. We compare WPE \cite{paper26} and the proposed AWEncoder in three downstream tasks with black-box access. The results are shown in Table I, where ``CE'' is short for clean encoder and ``WE'' is short for watermarked encoder. The fifth column shows the classification accuracy on the downstream task before and after watermarking. The last column gives $T_\text{cls}$ before/after watermarking. It is inferred that although there is a slight decrease in classification accuracy for the downstream task after watermarking, by assessing the difference between $T_\text{cls}$ for CE and WE, AWEncoder is more discriminative. 

In white-box scenario, we verify the ownership by contrasting the embedding similarity of clean images and watermarked images. Table II presents that the similarity score of WE is much lower than that of CE, which indicates that the similarity score can be used for reliable verification. When the watermark is generated with an incorrect key image (``dog'' in Fig. 1 is used in our experiment), $T_\text{sim}$ is much higher than the correct one. It indicates that AWEncoder has high security.

\subsection{Uniqueness}
To evaluate uniqueness, we generate forged watermarks by different methods including replacing \emph{Plane} with \emph{Dog} as the key image (see Fig. 1), changing the value of $\epsilon$, and replacing the pre-trained encoder based on ResNet-18 with a surrogate pre-trained encoder based on ResNet-50 \cite{paper34}. Table III shows the results, in which the similarity score and the classification score of the incorrect watermark are much higher than that of the correct one. It indicates that the proposed method provides superior performance on watermarking uniqueness. 

\subsection{Robustness}
In practice, adversaries may perform removal attacks such as fine-tuning and pruning to erase the watermark. To quantify the robustness of AWEncoder, we consider two common removal methods, i.e., fine-tuning all layers (FTAL) and retraining all layers (RTAL). FTAL fine-tunes the entire encoder with the training dataset and RTAL retrains the entire encoder with the downstream training dataset. In addition, we remove the parameters with the smallest L1-norms to prune the encoder. We also verify the robustness of AWEncoder in both white-box and black-box settings. The results in Table IV and Table V demonstrate that in white-box setting, pruning does not affect the watermarked encoder. Although fine-tuning increases $T_\text{sim}$ to a certain extent, AWEncoder can still effectively verify the ownership by applying a suitable threshold. In brief summary, AWEncoder is capable of resisting common attacks.

For black-box scenario, we compare AWEncoder with WPE by applying different attacks. The results are shown in Table VI and Table VII. It can be easily inferred that both fine-tuning and pruning will compromise the watermarking performance, however, AWEncoder is more robust than WPE against the removal attacks, which verifies the superiority of AWEncoder. 

\begin{table}[!t]
\centering
\caption{Robustness in black-box verification against fine-tuning.}
\label{table6}
\resizebox{\columnwidth}{!}{%
\begin{tabular}{c|c|c|ccccc|ccccc}
\hline\hline
\multirow{3}{*}{\begin{tabular}[c]{@{}c@{}}Fine-\\ tuning\end{tabular}} & \multirow{3}{*}{\begin{tabular}[c]{@{}c@{}}Downstream\\ dataset\end{tabular}} & \multirow{3}{*}{Method} & \multicolumn{5}{c|}{SimCLR} & \multicolumn{5}{c}{MoCo v2} \\ \cline{4-13} 
 &  &  & \multicolumn{2}{c|}{\begin{tabular}[c]{@{}c@{}}Accuracy\\ CE~/~WE\end{tabular}} & \multicolumn{3}{c|}{\begin{tabular}[c]{@{}c@{}}$T_\text{cls}$\\ CE~/~WE~/~$|$CE-WE$|$\end{tabular}} & \multicolumn{2}{c|}{\begin{tabular}[c]{@{}c@{}}Accuracy\\ CE~/~WE\end{tabular}} & \multicolumn{3}{c}{\begin{tabular}[c]{@{}c@{}}$T_\text{cls}$\\ CE~/~WE~/~$|$CE-WE$|$\end{tabular}} \\ \hline
\multirow{6}{*}{FTAL} & \multirow{2}{*}{ImageNet} & WPE & 74.8\% & \multicolumn{1}{c|}{74.3\%} & 22.3\% & 74.0\% & 51.7\% & 75.2\% & \multicolumn{1}{c|}{74.6\%} & 17.5\% & 76.4\% & 58.9\% \\
 &  & AWEncoder & 74.8\% & \multicolumn{1}{c|}{72.1\%} & 69.4\% & \textbf{7.6\%} & \textbf{61.8\%} & 75.2\% & \multicolumn{1}{c|}{73.1\%} & 69.5\% & \textbf{8.7\%} & \textbf{60.8\%} \\ \cline{2-13} 
 & \multirow{2}{*}{GTSRB} & WPE & 65.7\% & \multicolumn{1}{c|}{66.2\%} & 32.8\% & 61.7\% & 28.9\% & 88.6\% & \multicolumn{1}{c|}{86.5\%} & 15.5\% & 57.0\% & 41.5\% \\
 &  & AWEncoder & 65.7\% & \multicolumn{1}{c|}{63.2\%} & 76.1\% & \textbf{17.6\%} & \textbf{58.5\%} & 88.6\% & \multicolumn{1}{c|}{85.5\%} & 81.3\% & \textbf{16.6\%} & \textbf{64.7\%} \\ \cline{2-13} 
 & \multirow{2}{*}{STL-10} & WPE & 63.4\% & \multicolumn{1}{c|}{62.7\%} & 31.0\% & 86.7\% & 55.7\% & 64.6\% & \multicolumn{1}{c|}{64.6\%} & 15.6\% & 69.0\% & 53.4\% \\
 &  & AWEncoder & 63.4\% & \multicolumn{1}{c|}{60.8\%} & 85.4\% & \textbf{10.7\%} & \textbf{74.7\%} & 64.6\% & \multicolumn{1}{c|}{64.4\%} & 80.5\% & \textbf{20.5\%} & \textbf{60.0\%} \\ \hline
\multirow{6}{*}{RTAL} & \multirow{2}{*}{ImageNet} & WPE & 94.5\% & \multicolumn{1}{c|}{94.3\%} & 21.6\% & 60.5\% & 38.9\% & 96.0\% & \multicolumn{1}{c|}{96.2\%} & 22.5\% & 61.8\% & 39.3\% \\
 &  & AWEncoder & 94.5\% & \multicolumn{1}{c|}{92.7\%} & 60.2\% & \textbf{18.6\%} & \textbf{41.6\%} & 96.0\% & \multicolumn{1}{c|}{95.1\%} & 63.5\% & \textbf{23.9\%} & \textbf{39.6\%} \\ \cline{2-13} 
 & \multirow{2}{*}{GTSRB} & WPE & 98.5\% & \multicolumn{1}{c|}{98.9\%} & 29.7\% & 55.0\% & 25.3\% & 98.1\% & \multicolumn{1}{c|}{97.3\%} & 16.3\% & 48.0\% & 31.7\% \\
 &  & AWEncoder & 98.5\% & \multicolumn{1}{c|}{97.6\%} & 62.4\% & \textbf{21.6\%} & \textbf{40.8\%} & 98.1\% & \multicolumn{1}{c|}{97.1\%} & 64.8\% & \textbf{18.3\%} & \textbf{46.5\%} \\ \cline{2-13} 
 & \multirow{2}{*}{STL-10} & WPE & 83.1\% & \multicolumn{1}{c|}{82.2\%} & 24.6\% & 50.3\% & 25.7\% & 87.8\% & \multicolumn{1}{c|}{87.5\%} & 17.2\% & 41.7\% & 24.5\% \\
 &  & AWEncoder & 83.1\% & \multicolumn{1}{c|}{82.7\%} & 65.9\% & \textbf{20.0\%} & \textbf{45.9\%} & 87.8\% & \multicolumn{1}{c|}{85.2\%} & 61.3\% & \textbf{23.4\%} & \textbf{37.9\%} \\ \hline\hline
\end{tabular}%
}
\end{table}

\begin{table}[!t]
\centering
\caption{Robustness in black-box verification against pruning.}
\label{table7}
\resizebox{\columnwidth}{!}{%
\begin{tabular}{c|c|c|ccccc|ccccc}
\hline\hline
\multirow{3}{*}{\begin{tabular}[c]{@{}c@{}}Downstream\\ dataset\end{tabular}} & \multirow{3}{*}{\begin{tabular}[c]{@{}c@{}}Pruning\\ ratio\end{tabular}} & \multirow{3}{*}{Method} & \multicolumn{5}{c|}{SimCLR} & \multicolumn{5}{c}{MoCo v2} \\ \cline{4-13} 
 &  &  & \multicolumn{2}{c|}{\begin{tabular}[c]{@{}c@{}}Accuracy\\ CE~/~WE\end{tabular}} & \multicolumn{3}{c|}{\begin{tabular}[c]{@{}c@{}}$T_\text{cls}$\\ CE~/~WE~/~$|$CE-WE$|$\end{tabular}} & \multicolumn{2}{c|}{\begin{tabular}[c]{@{}c@{}}Accuracy\\ CE~/~WE\end{tabular}} & \multicolumn{3}{c}{\begin{tabular}[c]{@{}c@{}}$T_\text{cls}$\\ CE~/~WE~/~$|$CE-WE$|$\end{tabular}} \\ \hline
\multirow{8}{*}{GTSRB} & \multirow{2}{*}{0.2} & WPE & 79.8\% & \multicolumn{1}{c|}{80.3\%} & 30.8\% & 84.5\% & 53.7\% & 82.5\% & \multicolumn{1}{c|}{80.5\%} & 17.3\% & 80.4\% & 63.1\% \\
 &  & AWEncoder & 79.8\% & \multicolumn{1}{c|}{78.0\%} & 80.1\% & \textbf{14.6\%} & \textbf{65.5\%} & 82.5\% & \multicolumn{1}{c|}{81.6\%} & 84.2\% & \textbf{15.8\%} & \textbf{68.4\%} \\ \cline{2-13} 
 & \multirow{2}{*}{0.4} & WPE & 77.9\% & \multicolumn{1}{c|}{77.2\%} & 29.6\% & 79.2\% & 49.6\% & 79.1\% & \multicolumn{1}{c|}{78.8\%} & 19.6\% & 72.8\% & 53.2\% \\
 &  & AWEncoder & 77.9\% & \multicolumn{1}{c|}{76.3\%} & 80.5\% & \textbf{19.2\%} & \textbf{61.3\%} & 79.1\% & \multicolumn{1}{c|}{77.5\%} & 82.1\% & \textbf{19.4\%} & \textbf{62.7\%} \\ \cline{2-13} 
 & \multirow{2}{*}{0.6} & WPE & 71.2\% & \multicolumn{1}{c|}{70.5\%} & 27.9\% & 68.0\% & 40.1\% & 73.1\% & \multicolumn{1}{c|}{72.1\%} & 18.9\% & 62.6\% & 43.7\% \\
 &  & AWEncoder & 71.2\% & \multicolumn{1}{c|}{68.5\%} & 79.5\% & \textbf{20.4\%} & \textbf{59.1\%} & 73.1\% & \multicolumn{1}{c|}{70.0\%} & 77.6\% & \textbf{22.4\%} & \textbf{55.2\%} \\ \cline{2-13} 
 & \multirow{2}{*}{0.8} & WPE & 64.9\% & \multicolumn{1}{c|}{64.8\%} & 27.5\% & 65.7\% & 38.2\% & 65.4\% & \multicolumn{1}{c|}{65.7\%} & 17.0\% & 52.1\% & 35.1\% \\
 &  & AWEncoder & 64.9\% & \multicolumn{1}{c|}{63.8\%} & 77.9\% & \textbf{21.6\%} & \textbf{56.3\%} & 65.4\% & \multicolumn{1}{c|}{62.6\%} & 75.4\% & \textbf{24.0\%} & \textbf{51.4\%} \\ \hline\hline
\end{tabular}%
}
\end{table}

\section{Conclusion}
In this paper, we propose AWEncoder, an effective copyright protection technique for the pre-trained encoder in contrastive learning, which can be not only applied in both white-box and black-box scenarios but also transferred to several downstream tasks. Compared to the existing encoder watermarking, AWEncoder significantly improves the effectiveness and robustness. In the future, we will extend the proposed method to federated learning and ensemble learning, which are another two popular learning strategies widely applied in deep learning.

\section*{References}

\def\refname{}

\end{document}